\documentclass[10pt,twocolumn,letterpaper]{article}

\usepackage{cvpr}              

\usepackage{graphicx}
\usepackage{amsmath}
\usepackage{amssymb}
\usepackage{booktabs}

\usepackage[accsupp]{axessibility}

\usepackage{fancyhdr}

\usepackage{epsfig}

\usepackage{times}
\usepackage{mathtools}
\usepackage{url}
\usepackage{enumitem}

\usepackage{adjustbox}

\usepackage{subcaption}

\usepackage{multirow}
\usepackage[table,xcdraw]{xcolor}
\usepackage{comment}

\usepackage[pagebackref,breaklinks,colorlinks]{hyperref}

\usepackage[capitalize]{cleveref}
\crefname{section}{Sec.}{Secs.}
\Crefname{section}{Section}{Sections}
\Crefname{table}{Table}{Tables}
\crefname{table}{Tab.}{Tabs.}

\begin{document}

\title{Privacy-friendly Synthetic Data for the Development of Face Morphing Attack Detectors}

\author{
Naser Damer$^{1,2}$,  C\'{e}sar Augusto Fontanillo L\'{o}pez$^{3}$, Meiling Fang$^{1,2}$, No\'{e}mie Spiller$^{1}$ \\ Minh Vu Pham$^{1}$, Fadi Boutros$^{1,2}$\\
$^{1}$Fraunhofer Institute for Computer Graphics Research IGD,
Darmstadt, Germany\\
$^{2}$Department of Computer Science, TU Darmstadt,
Darmstadt, Germany\\
$^{3}$Centre for IT \& IP Law, KU Leuven, Leuven, Belgium\\
 Email: {naser.damer@igd.fraunhofer.de}
}

\maketitle

\begin{abstract}
\vspace{-2mm}
The main question this work aims at answering is: "can morphing attack detection (MAD) solutions be successfully developed based on synthetic data?".
Towards that, this work introduces the first synthetic-based MAD development dataset, namely the Synthetic Morphing Attack Detection Development dataset (SMDD).
This dataset is utilized successfully to train three MAD backbones where it proved to lead to high MAD performance, even on completely unknown attack types.
Additionally, an essential aspect of this work is the detailed legal analyses of the challenges of using and sharing real biometric data, rendering our proposed SMDD dataset extremely essential.
The SMDD dataset, consisting of 30,000 attack and 50,000 bona fide samples, is publicly available for research purposes \footnote{\url{https://github.com/naserdamer/SMDD-Synthetic-Face-Morphing-Attack-Detection-Development-dataset}}.

\end{abstract}

\vspace{-2mm}

\section{Introduction}
\vspace{-1mm}
Face recognition (FR) systems, despite their high accuracy \cite{DBLP:conf/cvpr/DengGXZ19} and acceptability \cite{Jain:1998:BPI:552539}, are vulnerable to many attacks \cite{DBLP:journals/cviu/MassoliCAF21,DBLP:conf/bmvc/DamerD16,DBLP:conf/btas/DamerWBBT0K18}, one of these is the face morphing attack (MA) \cite{DBLP:conf/icb/FerraraFM14}.
Ferrara et al. \cite{DBLP:conf/icb/FerraraFM14} analyzed face MAs by showing that an attack face image can be, automatically and by human experts, matched to more than one person. If MAs targeted travel or identity documents, they would enable multiple subjects to be verified to one document. This faulty link between a document and an identity can lead to illegal activities, including human trafficking, illegal immigration, and questionable financial transactions. 

Such attacks can, to a certain degree, be detected using MAD solutions \cite{NIST_Morph}.
MAD solutions typically require to be trained to differentiate between the two classes, MAs and bona fide (BF) (i.e. not attack) face images.
This training requires data containing samples of these classes to be used and possibly shared for MAD research and development by different persons and institutions.
Such data is, till now, based on real biometric data, limited in diversity and quantity, and raises ethical and legal challenges on its use and distribution, as will be discussed in detail in this paper.
This motivates our pursuit of using synthetic-based data for the development of MADs.
The pursue is based on the fact that MAD solutions are trained to detect the changes that the morphing process introduces to the image and makes the MA "different" in comparison to BF samples.
This detectable difference, therefore, might be detectable as well between synthetic images before and after undergoing a morphing process.

To address the need for large-scale MAD development datasets along with the legal and ethical challenges of using/reusing/sharing real biometric data, we raise the question of "can MAD solutions be successfully developed based on synthetic data?". 
As a first step towards our goal, we provide detailed and comprehensive legal analyses on the challenges raised by using real biometric data, thus motivating the move towards synthetic data in biometric solutions development.
We then present in this work our publicly available and synthetic-based SMDD dataset containing 30,000 MA and 50,000 BF samples.
By extensive experiments on three diverse MAD backbones and the evaluation over different MA variations, we can state that the MADs trained on our synthetic SMDD data did lead to stable MAD performances even on MAs created by unknown (not included in the training) morphing techniques.


\begin{figure*}
    \centering
    \includegraphics[width=0.850\linewidth]{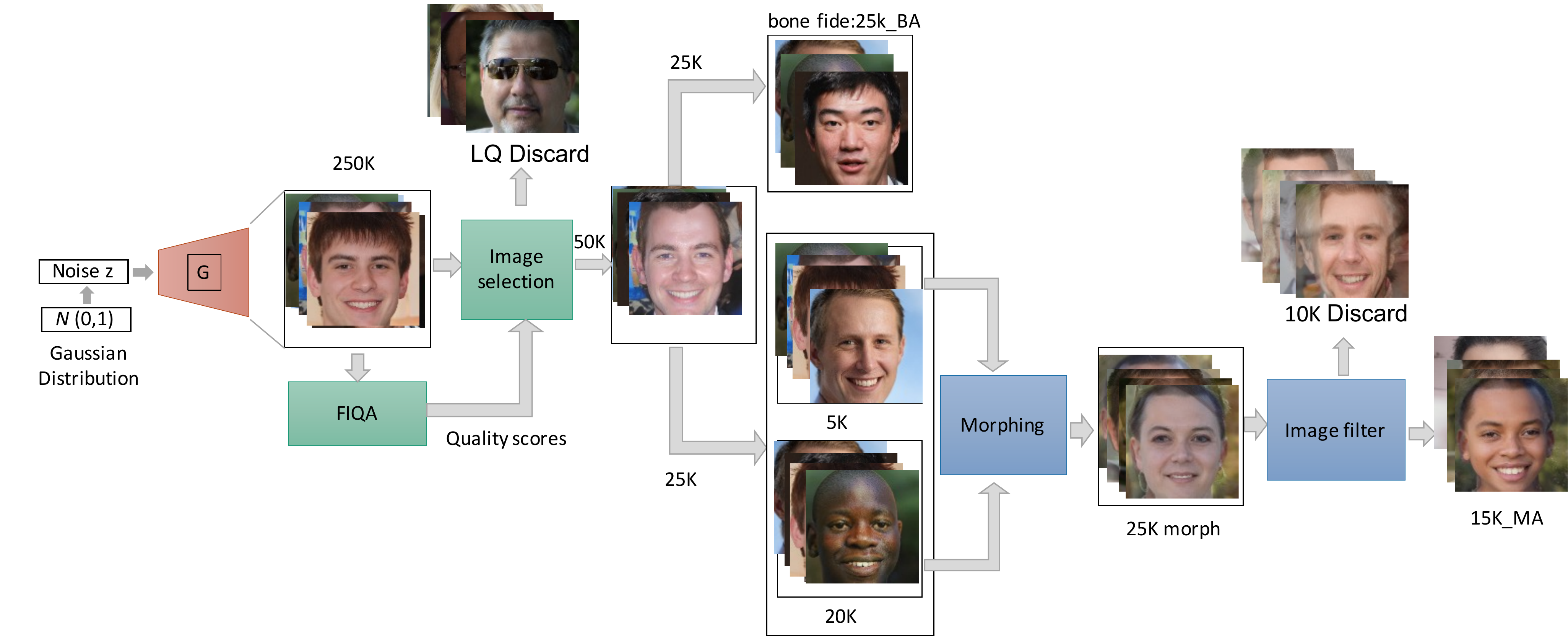}
    \vspace{-2mm}
    \caption{Workflow of the SMDD dataset creation, starting from synthetic faces to the final MA and BF sets.}
    \label{fig:flow}
    \vspace{-3mm}
\end{figure*}

\vspace{-2mm}
\section{Related work}
\vspace{-2mm}


The creation of face MA datasets started with the initial works that uncovered the attack possibility \cite{DBLP:conf/icb/FerraraFM14}.
This was followed by a set of datasets that were created to train and evaluate MADs \cite{DBLP:conf/btas/RaghavendraRB16,DBLP:conf/icb/RaghavendraRVB17,DBLP:conf/cvpr/RaghavendraRVB17a} and to further study the vulnerability of FR to MAs \cite{DBLP:conf/iwbf/ScherhagRRGRB17, DBLP:books/sp/16/FerraraFM16}. Most of such datasets created MAs based on the interpolation of facial landmarks, whether detected manually or automatically \cite{DBLP:books/sp/16/FerraraFM16,DBLP:conf/btas/RaghavendraRB16,DBLP:journals/tifs/ScherhagRMB20}.
Another face morphing strategy based on generative adversarial networks (GAN) was proposed in \cite{DBLP:conf/btas/DamerS0K18}. This was later enhanced with a cascaded refinement network to possess higher perceived quality \cite{DBLP:conf/btas/DamerBSKK19}.
Using more advanced GAN architectures, \cite{MIPGAN} and \cite{DBLP:conf/iwbf/VenkateshZRRDB20} produced more realistic GAN generated morphs. 
The post-processing of landmark-based face morphs in a GAN architecture was later proposed in \cite{ReGenM}.
Some of the created datasets focused on creating versions with certain variations such as morphing pair selection protocols \cite{DBLP:conf/icb/DamerSZWTKK19,DBLP:conf/fusion/DamerZWSKK19}, print and scan operations \cite{PW_MAD,https://doi.org/10.1049/bme2.12021}, or variations in landmark based morphing \cite{DBLP:journals/tifs/ScherhagRMB20,DBLP:journals/tbbis/ScherhagDRBU19,Sarkar2020}.

Most of of these existing datasets only included up to a thousand MAs per morphing strategy \cite{DBLP:books/sp/16/FerraraFM16,DBLP:conf/icb/FerraraFM14,DBLP:conf/btas/RaghavendraRB16,DBLP:conf/iwbf/ScherhagRRGRB17,DBLP:conf/cvpr/RaghavendraRVB17a,DBLP:conf/iwbf/Gomez-BarreroRS17,DBLP:journals/tifs/FerraraFM18,DBLP:conf/btas/DamerS0K18,DBLP:conf/btas/DamerBSKK19,PW_MAD,DBLP:journals/tbbis/ScherhagDRBU19,https://doi.org/10.1049/bme2.12021,DBLP:conf/biosig/FerraraFM19,DBLP:conf/btas/DamerGZKK19,Sarkar2020,DBLP:conf/dagm/DamerBWBTBK18}, which might be insufficient to train an over-parameterized deep neural network and had very limited variation to create a generalizable MAD.
Larger datasets were created for this reason with over a thousand morphs per morphing strategy, but also were restricted to under 4 thousand morphs \cite{DBLP:conf/icb/RaghavendraRVB17,DBLP:conf/cvip/RamachandraVRB18,DBLP:journals/tifs/ScherhagRMB20,DBLP:conf/iwbf/VenkateshZRRDB20,MIPGAN,ReGenM}, which is still hundreds of folds smaller in comparisons to datasets used for common FR \cite{DBLP:journals/corr/YiLLL14a,guo2016ms} or spoof detection training \cite{DBLP:conf/eccv/ZhangYLYYSL20}.

From these discussed MA datasets, only a few are available for MAD development research. To the best of our knowledge, the available datasets are the MorGan and associated LMA sets with 1000 MAs, and their enhanced versions \cite{DBLP:conf/btas/DamerS0K18,DBLP:conf/btas/DamerBSKK19}, the limited print and scan LMA-DRD dataset with only 276 attacks \cite{PW_MAD}, and the recently released FRLL-Morphs, FERET-Morphs, and FRGC-Morphs, each with under or around one thousand attacks per most morph types \cite{Sarkar2020}.
FRLL-Morphs, FERET-Morphs, and FRGC-Morphs datasets are not presented as an MAD training or evaluation set, and they aim at enabling the evaluation of FR vulnerability to MAs.

The fact of the limited size of the existing face MA datasets, the privacy issues that limits their public availability, and the ethical concerns on sharing and reusing biometric information of individuals, stresses the need for MAD development datasets that are of a large-scale, privacy-friendly, and can be shared in the research community. This gap is explored here by the proposal of the SMDD dataset.







\vspace{-2mm}
\section{The legal need for synthetic-based solutions}
\vspace{-1mm}
Biometric data processing is a legally challenging undertaking. Since the entry into force of the General Data Protection Regulation (GDPR) \cite{eu-2016/679},  biometric data has acquired legal recognition of a special category of of personal data while being subject to a strict data protection regime \cite{art9gdpr}.
According to this body of law, the processing of special categories of personal data, including biometric data, requires the highest standards of protection in view of the fundamental rights and freedoms of individuals. 
This translates into onerous legal obligations for biometric data controllers, including, among others, (i) the necessity to adhere to one of the exemptions of biometric data processing cumulatively with the choice of a suitable legal basis \cite{art9_2gdpr};  (ii) the need to comply with the national law of the European Union (EU) Member State relating to the processing of biometric data, where applicable \cite{art9_4gdpr};  (iii) the maintenance of processing records \cite{art30gdpr};  and (iv) the preparation of a data protection impact assessment and the appointment of a DPO, where applicable \cite{art35_3gdpr,art37_1gdpr}.  Moreover, depending on the circumstances and the intended purposes of the biometric data processing, data controllers would also be required to (v) comply with the mandatory obligations relating to automated individual decision-making, where applicable \cite{art22_4gdpr};  (vi) assume the loss of any possibility of exemption with respect to the appointment of a representative in the EU for controllers and processors located outside the EU \cite{art27_2gdpr},  and, (vii) where biometric data is processed for purposes other than those initially foreseen, consider the merits and demerits of their compatibility \cite{art6_4gdpr}.  These legal obligations are, in common practice, supplemented by other ethical requirements, such as the approval of the biometric data processing by the respective ethics committee or competent authority prior to its onset, particularly in the academic sector \footnote{  For instance, copies of opinions and/or approvals by ethics committees and/or competent authorities for biometric data collections are common requirements under EU-funded projects. In certain cases, ethics advisors are also recruited to ensure the responsible, morally acceptable and legally compliant research oversight. E.g. image Manipulation Attack Resolving Solutions (iMARS) Project, 
\url{https://imars-project.eu/}}.

Given the exigent legal and ethical framework, the carrying out of biometric data processing can become a tedious, time-consuming, and resource-intense task. More particularly, biometric data sharing and reuse are deemed as some of the most challenging situations faced by data controllers due to the bureaucracy imposed by data protection law. Compliance with the GDPR would require, in these cases, the approach to the data subjects to request their explicit consent where no other exception and legal basis can provide for a lawful and fair sharing or reuse of the biometric data, and where the 'compatibility test' \cite{art6_4gdpr}  cannot offer a legally viable solution. For instance, where the reuse of biometric data cannot be based on the scientific research exception \cite{art9_2_jgdpr} coupled with the need to fulfill a task performed for reasons of public interest \cite{art6_1egdpr},  and the 'compatibility test' does not offer a positive result in light of, inter alia, the sensitivity of biometric data \cite{art6_4cgdpr}, the principle of fairness \cite{art5_1agdpr} would require, considering the inadequacy of the other available exceptions \cite{art9_2b_igdpr} and legal bases \cite{art6_1bcdfgdpr}, recourse to explicit consent \cite{art9_2agdpr} cumulatively with the consent \cite{art6_1agdpr} of the data subject as the appropriate exception and legal basis, respectively, for the reuse of biometric data. Furthermore, the exercise of data subject's rights may, in some instances, jeopardize these practices or even be unrealizable for the biometric data controller \footnote{Most prominently, the right to withdraw the consent of Art. 7(3) GDPR  \cite{art7_3bcdfgdpr}  and the right to erasure of Art. 17 GDPR \cite{art17bcdfgdpr}.}.  In this vein, compliance with the withdrawal of consent may, depending on the number of requests, allegedly undermine the subsequent reuse of biometric data. This may, in turn, impact the high investments that public and private institutions make for the furtherance of biometrics. Similarly, the exercise of the right of erasure by the data subject after the sharing or publication of its biometric data may not be possible or may not be effectively enforced by the data controller once the biometric data have left its sphere of control. In a global context of competition and European cohesion characterized by the pressing need to institute growing databases rich in biometric data to advance the state-of-the-art in scientific and industrial applications as well as data governance \cite{frattini.comissionsPolicyPriorities,europeanComission.PrioritySocietalChallenges,europeanParliament.EuropeanParliamentaryResearchService,COM/2020/66final,COM/2020/65final},  it is debatable how the free flow of information can be achieved in the EU while ensuring strict compliance with the right to the protection of personal data. More importantly, it may be questionable how public and private actors may comply with the stringent regime applicable to biometric data without compromising or seriously impairing the achievement of biometric data sharing or reuse. Based on these assumptions, the legally admissible regime for the processing of biometric data could end up being a practically unbearable regime for data controllers, primarily where databases comprise the biometric data of a large number of data subjects. 

To overcome the dilemma of 'protection of personal data - free flow of information', data controllers recurrently seek to anonymize personal data. However, since the threshold for anonymization is being set very high by the European legislator \cite{Recital.26.GDPR},  notorious expenditures in terms of data transformation procedures and control mechanisms must be made by the data controllers to anonymize personal data.
In addition to this, the anonymization of personal data creates other undesired situations, such as the trade-off between privacy and utility \cite{DBLP:journals/tifs/MedenRTDKSRPS21}. It is for these reasons that more effective data-driven approaches have been recently explored to cope with the pitfalls of current anonymization techniques \cite{4497436,dwork.differentialPrivacy,Lindell_Pinkas_2009}.  Among them, the use of synthetic data is gaining increasing attention in the technical and legal community due to its towering advantages over conventional anonymization techniques \cite{9138552}.  
It is argued by some that the particular properties of synthetic data will offer an alternative to private and public actors for research and knowledge transfer, while others have doubts \cite{277172}. Advantages of synthetic data include that data controllers may circumnavigate the legal constraints of biometric data processing while maintaining utility, thus permitting the alignment of innovation and law \cite{emam2020practical}, the main motivation behind this work.

\begin{figure}[ht!]
    \centering 
    \includegraphics[width=0.7\linewidth]{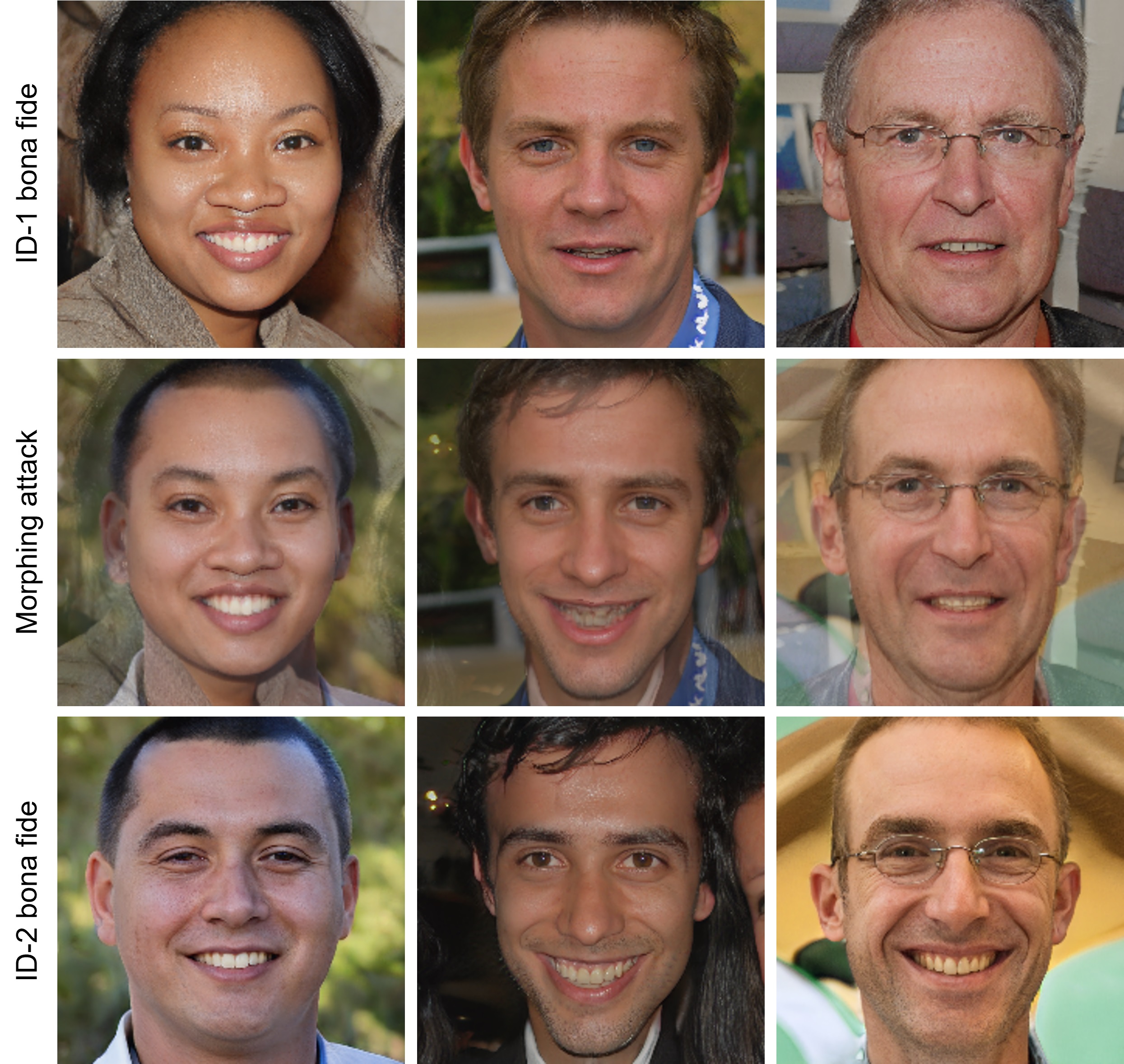}
    \vspace{-2mm}
    \caption{Samples of the synthetic SMDD dataset. The top and bottom rows are BF samples. The middle row is samples of the attacks created from the corresponding images above and below the morphed image.}
    \label{fig:smdd_samp}
    \vspace{-3mm}
\end{figure}

\vspace{-2mm}
\section{The SMDD dataset}
\vspace{-1mm}
This section presents our Synthetic Morphing Attack Detection Development dataset (SMDD). This is the first synthetic-based biometric attack detection dataset to be proposed and released to the research community. 
The overview of the dataset creation process is presented in Fig. \ref{fig:flow}.
To start, we created 500k synthetic images using the official open source implementation \footnote{\url{https://github.com/NVlabs/stylegan2-ada}} of StyleGAN2-ADA \cite{DBLP:conf/nips/KarrasAHLLA20}  trained on Flickr-Faces-HQ Dataset (FFHQ) dataset \cite{DBLP:conf/cvpr/KarrasLA19}.
These images are randomly generated based on a Gaussian noise Z, drown without repetition 500k times, from a normal distribution. 
From each latent vector Z, the pretrained generator produces a synthetic face data sample.
These data samples are split randomly into 250k samples that will be the basis of the training data, and a 250k samples that will be the basis of the evaluation data.
The main goal of the SMDD dataset is to provide a training dataset for MADs, however, a disjoint testing dataset is created to investigate the realistic deployability of such synthetic datasets as an evaluation benchmark as well.
For each of these two sets, the same process described in Fig. \ref{fig:flow} is performed. 
The 250k samples are filtered automatically by using the 50k samples with the highest face image quality, calculated by the CR-FIQA (S) quality assessment approach \cite{DBLP:journals/corr/abs-2112-06592}. 
The CR-FIQA (S) quality assessment approach \cite{DBLP:journals/corr/abs-2112-06592} reflects the utility of the sample for FR, as defined in ISO/IEC 29794-1 \cite{ISOIEC29794-1}, and thus reflects extreme issues that effect FR performance (e.g. extreme poses or occlusions) in a low quality score.
This automatic filtering removed most of the images with extreme non-frontal poses and largely occluded (sunglasses and generation artifacts). 
These filtered 50k images are randomly split into two 25k parts, the first will be considered the BF samples, namely the 25k-BF, and will not go into a morphing process.
The second part will be used to create the face MAs.
5k of these images are chosen randomly as the key morphing images, each is paired with 5 images chosen randomly from the 20k left images of the 25k images aimed to create the attacks. 
This results in 25k morphing pairs.
The morphing pairs are selected randomly to allow for a larger variety of similarities between the pairs.
Although high similarity pairs create stronger attacks (more successful at attacking FR systems) \cite{DBLP:conf/icb/DamerSZWTKK19}, having variation in the pair similarity in the MAD training data has shown to lead to more generalizable MADs \cite{DBLP:conf/icb/DamerSZWTKK19}.
Given that training MADs is the main reason behind the creation of SMDD, we chose the random selection of morphing pairs.
Each of the pairs is morphed using the widely used \cite{DBLP:journals/tifs/ScherhagRMB20,DBLP:conf/btas/DamerS0K18,Sarkar2020} OpenCV/dlib  morphing algorithm \cite{openCVmorph} based
on the “Face Morph Using OpenCV” tutorial\footnote{\url{http://www.learnopencv.com/face-morph-using-opencv-cpp-python/}} using the Dlib \cite{dlib09} implementation of \cite{DBLP:conf/cvpr/KazemiS14} landmark detector.
This morphing tool proved to produce MAs stronger than many other comparable landmark-based morphing tools \cite{Sarkar2020}.
The facial landmarks are detected in each of the morphed images \cite{DBLP:conf/cvpr/KazemiS14}.
Then, a Delaunay triangulation is performed on each image base don these detected landmarks.
Based on calculated affine transforms, the triangles are wrapped with a wrapping factor and the images are blended with a blending factor (also known as a morphing factor).
We follow the recommended factor, leading to the most successful attack setups in \cite{DBLP:conf/biosig/FerraraFM19} by setting the blending factor to 0.5 and choosing the warping factor randomly from the range [0.0,1.0].
Examples of the morphed images in our SMDD, along with the synthetic images that created the morphs, are presented in Fig. \ref{fig:smdd_samp}, where one can note the realistic appearance of the MA and BF samples.
The resulting 25k MA images are filtered again by removing images with common automatic landmark-based morphing artifacts (black regions on the mouth area) and by manually inspecting the morphed images and disregarding ones with very strong morphing artifacts.
This results in reducing the 25k morphed images down to the final set of 15k MAs, namely the 15k-MA.
The same process is duplicated for the disjoint training and evaluation sets resulting in a training set of 25k BF and 15k MA samples (the T-25k-BF and T-15k-MA), and the evaluation set of 25k BF and 15k MA samples (the E-25k-BF and E-15k-MA). 
Examples of the BF and MA samples in the SMDD dataset is presented in Fig. \ref{fig:smdd_samp}.

\begin{figure}
    \centering
     \begin{subfigure}[b]{0.18\textwidth}
         \centering
         \includegraphics[width=0.47\textwidth]{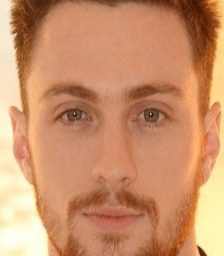}
         \includegraphics[width=0.47\textwidth]{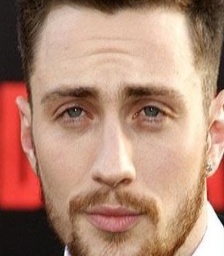}
         \caption{LMA-DRD(D)}
         \label{fig:bf_drd_d}
     \end{subfigure}
     \begin{subfigure}[b]{0.18\textwidth}
         \centering
         \includegraphics[width=0.47\textwidth]{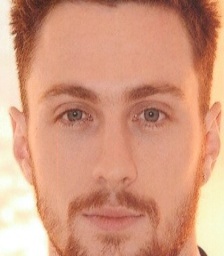}
         \includegraphics[width=0.47\textwidth]{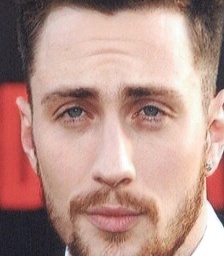}
         \caption{LMA-DRD(PS)}
         \label{fig:bf_drd_ps}
     \end{subfigure}
     \begin{subfigure}[b]{0.18\textwidth}
         \centering
         \includegraphics[width=0.47\textwidth]{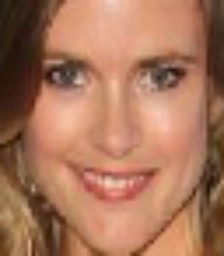}
         \includegraphics[width=0.47\textwidth]{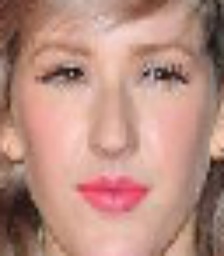}
         \caption{MorGAN}
         \label{fig:bf_morgan}
     \end{subfigure}
     \begin{subfigure}[b]{0.18\textwidth}
         \centering
         \includegraphics[width=0.47\textwidth]{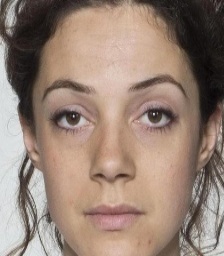}
         \includegraphics[width=0.47\textwidth]{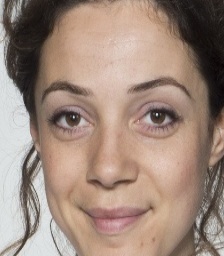}
         \caption{FRLL-Morphs}
         \label{fig:bf_FRLL}
     \end{subfigure}
    \vspace{-3mm}
    \caption{BF samples from the (a) LMA-DRD (D) \cite{PW_MAD}, (b) LMA-DRD (PS) \cite{PW_MAD}, (c) MorGAN-LMA/GAN \cite{DBLP:conf/btas/DamerS0K18} (MorGan-LMA and MorGan-GAN share bona fide sets), (d) FRLL-Morphs \cite{Sarkar2020}.}
    \label{fig:dat_Samples_bf}
    \vspace{-4mm}
\end{figure}

\begin{figure*}
    \centering
     \begin{subfigure}[b]{0.175\textwidth}
         \centering
         \includegraphics[width=0.47\textwidth]{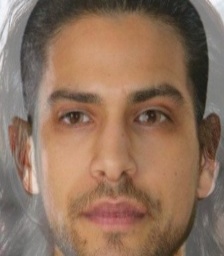}
         \includegraphics[width=0.47\textwidth]{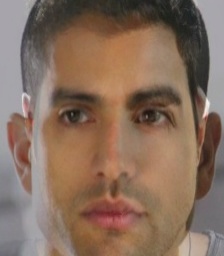}
         \caption{LMA-DRD(D)}
         \label{fig:bf_drd_d}
     \end{subfigure}
     \begin{subfigure}[b]{0.175\textwidth}
         \centering
         \includegraphics[width=0.47\textwidth]{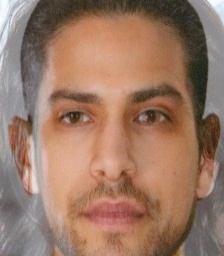}
         \includegraphics[width=0.47\textwidth]{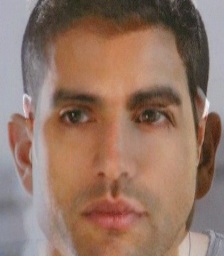}
         \caption{LMA-DRD(PS)}
         \label{fig:bf_drd_ps}
     \end{subfigure}
     \begin{subfigure}[b]{0.175\textwidth}
         \centering
         \includegraphics[width=0.47\textwidth]{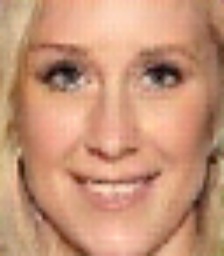}
         \includegraphics[width=0.47\textwidth]{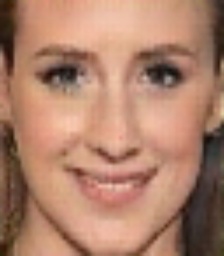}
         \caption{MorGAN-GAN}
         \label{fig:bf_morgan}
     \end{subfigure}
     \begin{subfigure}[b]{0.175\textwidth}
         \centering
         \includegraphics[width=0.47\textwidth]{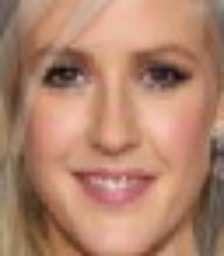}
         \includegraphics[width=0.47\textwidth]{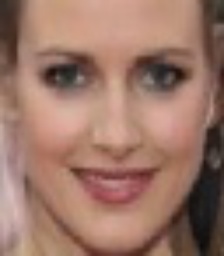}
         \caption{MorGAN-LMA}
         \label{fig:bf_morgan}
     \end{subfigure}
     \begin{subfigure}[b]{0.175\textwidth}
         \centering
         \includegraphics[width=0.47\textwidth]{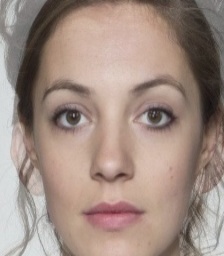}
         \includegraphics[width=0.47\textwidth]{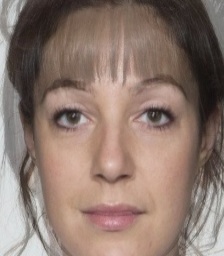}
         \caption{FRLL-OCV}
         \label{fig:bf_FRLL}
     \end{subfigure}
     \begin{subfigure}[b]{0.175\textwidth}
         \centering
         \includegraphics[width=0.47\textwidth]{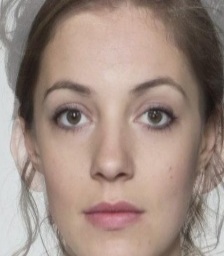}
         \includegraphics[width=0.47\textwidth]{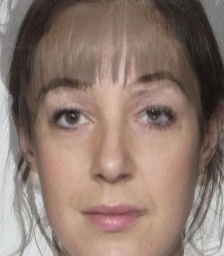}
         \caption{FRLL-FM}
         \label{fig:bf_FRLL}
     \end{subfigure}
     \begin{subfigure}[b]{0.175\textwidth}
         \centering
         \includegraphics[width=0.47\textwidth]{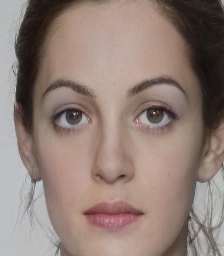}
         \includegraphics[width=0.47\textwidth]{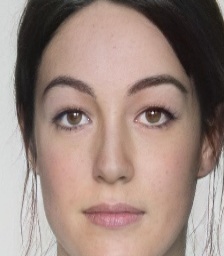}
         \caption{FRLL-SG}
         \label{fig:bf_FRLL}
     \end{subfigure}
     \begin{subfigure}[b]{0.175\textwidth}
         \centering
         \includegraphics[width=0.47\textwidth]{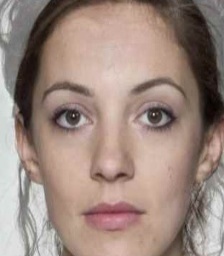}
         \includegraphics[width=0.47\textwidth]{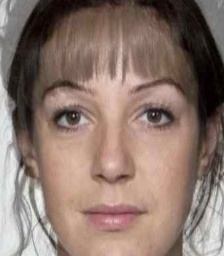}
         \caption{FRLL-WM}
         \label{fig:bf_FRLL}
     \end{subfigure}
     \begin{subfigure}[b]{0.175\textwidth}
         \centering
         \includegraphics[width=0.47\textwidth]{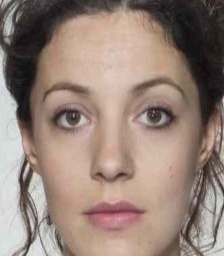}
         \includegraphics[width=0.47\textwidth]{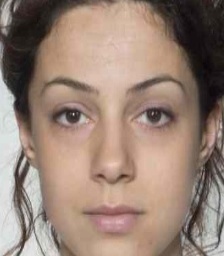}
         \caption{FRLL-AMSL}
         \label{fig:bf_FRLL}
     \end{subfigure}
    \vspace{-2.5mm}
    \caption{Morphing attack samples from the (a) LMA-DRD (D) \cite{PW_MAD}, (b) LMA-DRD (PS) \cite{PW_MAD}, (c) MorGAN-GAN \cite{DBLP:conf/btas/DamerS0K18}, (d) MorGAN-LMA \cite{DBLP:conf/btas/DamerS0K18}, and (e-i) the different attack types of the FRLL-Morphs \cite{Sarkar2020}.}
    \label{fig:dat_Samples_ma}
    \vspace{-4mm}
\end{figure*}

The complete dataset is available publicly for research purposes.
One limitations, and opportunity, of the SMDD dataset is that it only creates MAs using one morphing strategy. However, to enable extending such privacy-friendly dataset in the research community, both the images used to create the MAs and the exact morphing pairs list will be publicly available, where researchers can use them to build subsequent synthetic-based face morphing datasets.
\vspace{-3mm}
\section{MAD and experimental setup}
\vspace{-1mm}
To explore the deployability of the SMDD dataset as a development set for face MAD, we train three different MAD backbones using the SMDD training set. We evaluate the resulting MAD on a set of different MA datasets with variations including print and scan operations and a wide range of morphing techniques, including landmark-based and GAN-based ones. 
To compare the applicability of SMDD as an MAD development set, we compare the performance of the MADs trained on SMDD with the same set of MADs trained on four other datasets. This comparison is not direct as the synthetic nature of the SMDD dataset allows to create a much larger number of samples, unlike conventional real data, however, the performance comparison aims at putting the achieved performances in perspective. This section introduces technical details about the used MADs, the baseline training datsets, the test datasets, the experimental settings, and the MAD evaluation metrics.
\vspace{-1mm}
\subsection{Morphing attack detectors}
\vspace{-1mm}
To give a generalized look at the possibility of using SMDD for MAD, we use three diverse MAD backbones.

\noindent \textbf{Inception-MAD}: This baseline uses the Inception-v3 \cite{inception_v3} network architecture as its backbone. This architecture has been used successfully for MAD \cite{DBLP:conf/cvip/RamachandraVRB18,PW_MAD} and fake face detection approaches \cite{DBLP:conf/biosig/KhodabakhshRRWB18}. 
We report the results of a trained from scratch Inception-v3 model, named $\mathrm{Inception MAD}$. 
In the training phase, the binary cross-entropy loss function and Adam optimizer with a learning rate of $10^{-4}$ and a weight decay of $10^{-5}$ are used. The input of the network is of the size $229 \times 229 \times 3$. Moreover, early stopping technique with the patience of 20 epochs is used to avoid overfitting and for a fair comparison.

\noindent \textbf{MixFaceNet-MAD}: The second utilized backbone architecture is MixFaceNet \cite{DBLP:conf/icb/BoutrosDFKK21}, which is an extremely efficient architecture for FR. MixFaceNet possesses lower computation complexity (FLOPs) and high accuracy and was used successfully in face presentation attack detection \cite{DBLP:conf/fgr/FangBKD21}. Moreover, MixFaceNet architecture takes advantage of different sizes of convolutional kernels to capture different levels of face attack clues \cite{DBLP:conf/fgr/FangBKD21}. In our experiment, MixFaceNet, specifically MixFaceNet-S, is used as it represent extremely efficient network.
The input image size \cite{DBLP:conf/icb/BoutrosDFKK21} of MixFaceNet is modified to $224 \times 224 \times 3$ and a fully-connected layer is added after the embedding layer to perform the MAD decision. In the training phase, binary cross-entropy loss function, Adam optimizer, and early stopping are used with the same hyper-parameters with the training of Inception-MAD.


\noindent \textbf{PW-MAD:} As a complement to the binary label supervision, PW-MAD \cite{PW_MAD} utilizes pixel-wise map supervision \cite{deeppix_19}, i.e. a label of attack or BF for each image pixel. Considering the superior MAD performance obtained by the PW-MAD method \cite{PW_MAD}, we choose the PW-MAD \cite{PW_MAD} as our third MAD. The loss for training PW-MAD is the sum of the binary cross-entropy loss computed based on pixel-wise feature map and binary output following the definition in \cite{PW_MAD}. Our solution replaces the last sigmoid layer in \cite{PW_MAD} with a softmax layer which proved to enhance generalizability over unknown morphing techniques. The input of the network is of the size $224 \times 224 \times 3$. The hyper-parameters of Adam and early stopping technique is same to used in Inception-MAD and MixFaceNet-MAD.

It must be noted that all the MAD solutions used only the training part of the respective datasets for training, which is identity disjoint from all the testing sets.
All the images used for the training and testing are cropped using the MTCNN \cite{mtcnn} bounding box detection with 5\% extension of the width and height to include the whole face following \cite{PW_MAD} (examples in Fig. \ref{fig:dat_Samples_bf} and \ref{fig:dat_Samples_ma}) and all crops are resized to $224 \times 224$. 

\begin{table}
\centering
\resizebox{0.39\textwidth}{!}{
\begin{tabular}{|c|l|l|l|llll|}
\hline
\multirow{2}{*}{\rotatebox[origin=c]{90}{Test}}                         & \multicolumn{1}{c|}{\multirow{2}{*}{MAD}} & \multicolumn{1}{c|}{\multirow{2}{*}{Train Data}} & \multicolumn{1}{c|}{\multirow{2}{*}{EER (\%)}} & \multicolumn{4}{c|}{BPCER   (\%) @ APCER =}                                                                                                    \\ \cline{5-8} 
                                              & \multicolumn{1}{c|}{}                     & \multicolumn{1}{c|}{}                            & \multicolumn{1}{c|}{}                          & \multicolumn{1}{c|}{0.10\%}         & \multicolumn{1}{c|}{1.00\%}         & \multicolumn{1}{c|}{10.00\%}        & \multicolumn{1}{c|}{20.00\%} \\ \hline 
\multirow{15}{*}{\rotatebox[origin=c]{90}{FRLL-Morphs - OpenCV}}      & MixFacenet                                & LMA-DRD (D)                                      & 8.83                                           & \multicolumn{1}{l|}{82.84}          & \multicolumn{1}{l|}{34.80}          & \multicolumn{1}{l|}{8.82}           & 3.43                         \\ \cline{2-8} 
                                              & MixFacenet                                & LMA-DRD (PS)                                     & 13.22                                          & \multicolumn{1}{l|}{64.71}          & \multicolumn{1}{l|}{46.08}          & \multicolumn{1}{l|}{16.18}          & 7.84                         \\ \cline{2-8} 
                                              & MixFacenet                                & MorGAN-LMA                                       & 8.91                                           & \multicolumn{1}{l|}{\textbf{56.86}} & \multicolumn{1}{l|}{\textbf{17.16}} & \multicolumn{1}{l|}{8.33}           & 4.41                         \\
\hline
\end{tabular}}
\vspace{-2mm}
\caption{MAD performances on the different attacks of the FRLL-Morphs when each of the three MADs is trained on LMA-DRD (D), LMA-DRD (PS), MorGAN-GAN, MorGAN-GAN, or SMDD training data. The lowest error rates per MAD are in bold.}
\label{tab:res_unknown}
\vspace{-3mm}
\end{table}

\vspace{-1mm}
\subsection{Datasets}
\label{sec:data}
\vspace{-1mm}
Beside our SMDD, we utilize five other datasets, LMA-DRD (PS) \cite{PW_MAD}, LMA-DRD (D) \cite{PW_MAD}, MorGAN-LMA \cite{DBLP:conf/btas/DamerS0K18}, MorGAN-GAN \cite{DBLP:conf/btas/DamerS0K18}, and FRLL-Morphs \cite{Sarkar2020}. Of these, the first four contain identity-disjoint test and training sets, where the latter are used to train MAD and put the MAD trained on SMDD in perspective. The fifth contains only test data with five different morphing attack techniques and is used to evaluate the resulting MADs on unknown ( in terms of morphing technique and original dataset) attacks.

The \textbf{LMA-DRD (D) and LMA-DRD (PS)} are built on selected images  from the VGGFace2 \cite{DBLP:conf/fgr/CaoSXPZ18} dataset. These are selected to be frontal, of high resolution, and with neutral expression \cite{PW_MAD}. The morphing method is based on OpenCV morphing \cite{openCVmorph} and following parametrization in \cite{DBLP:conf/icb/RaghavendraRVB17}. BF images are also selected such that it follows the mentioned manual quality check. The dataset is split into two identical sets, one is digital and noted as LMA-DRD (D) and one is re-digitized (print and scan) and noted as LMA-DRD (PS). From both, we used the train and development sets for training (181 morphs and 241 BF images) and the identity-disjoint test set for testing (88 morphs and 123 BF images).

\begin{table}
\centering
\resizebox{0.39\textwidth}{!}{
\begin{tabular}{|c|l|l|l|llll|}
\hline
                                  & \multicolumn{1}{c|}{}                      & \multicolumn{1}{c|}{}                             & \multicolumn{1}{c|}{}                           & \multicolumn{4}{c|}{BPCER   (\%) @ APCER =}                                                                                                                                                  \\ \cline{5-8} 
\multirow{-2}{*}{\rotatebox[origin=c]{90}{Test}}            & \multicolumn{1}{c|}{\multirow{-2}{*}{MAD}} & \multicolumn{1}{c|}{\multirow{-2}{*}{Train Data}} & \multicolumn{1}{c|}{\multirow{-2}{*}{EER (\%)}} & \multicolumn{1}{c|}{0.10\%}                        & \multicolumn{1}{c|}{1.00\%}                        & \multicolumn{1}{c|}{10.00\%}                       & \multicolumn{1}{c|}{20.00\%}  \\ \hline
                                  & MixFacenet                                 & LMA-DRD (D)                                       & \cellcolor[HTML]{E7E6E6}15.68                   & \multicolumn{1}{l|}{\cellcolor[HTML]{E7E6E6}82.11} & \multicolumn{1}{l|}{\cellcolor[HTML]{E7E6E6}69.92} & \multicolumn{1}{l|}{\cellcolor[HTML]{E7E6E6}30.08} & \cellcolor[HTML]{E7E6E6}13.01 \\ \cline{2-8} 
                                  & MixFacenet                                 & LMA-DRD (PS)                                      & 18.03                                           & \multicolumn{1}{l|}{65.04}                         & \multicolumn{1}{l|}{\textbf{56.10}}                & \multicolumn{1}{l|}{\textbf{30.89}}                & 17.89                         \\ \cline{2-8} 
                                  & MixFacenet                                 & MorGAN-LMA                                        & \textbf{17.06}                                  & \multicolumn{1}{l|}{\textbf{56.91}}                & \multicolumn{1}{l|}{56.91}                         & \multicolumn{1}{l|}{32.52}                         & \textbf{16.26}                \\
\hline
\end{tabular}
}
\vspace{-2mm}
\caption{MAD performances on the LMA-DRD (D), LMA-DRD (PS), MorGAN-LMA, MorGAN-GAN, and SMDD test data, when each of the three MADs is trained on the train sets of the same five datasets. Shadowed rows represent testing on the same data source used for training (although identity-disjoint sets), for the rest of the rows, the lowest error rates per MAD are in bold.}
\vspace{-3mm}
\label{tab:res_known}
\end{table}

The \textbf{MorGAN-LMA and MorGAN-GAN} datasets \cite{DBLP:conf/btas/DamerS0K18} created MAs based on selected images from CelebA \cite{DBLP:conf/iccv/LiuLWT15}. The image selection was based on the face being frontal and the morphing pairs were selected so that they are the most similar pairs (based on OpenFace \cite{openface} FR solution) as described in \cite{DBLP:conf/iccv/LiuLWT15}. The morphing is either based on the OpenCV landmark-based morphing \cite{openCVmorph} or on the GAN-based solution presented in \cite{DBLP:conf/btas/DamerS0K18}, resulting in the MorGAN-LMA and MorGAN-GAN datasets, respectively. Each of these two sets contains identity-disjoint training and testing sets. 
Each of the training and testing sets contains 500 MA and 750 BF samples. One aspect that makes this dataset less realistic, but essential to study, is that the generated faces were relatively of low resolution of $64 \times 64$ pixels.

The \textbf{FRLL-Morphs} dataset \cite{Sarkar2020} is generated from the publicly available Face Research London Lab dataset \cite{DeBruine_FRLL}. The dataset include 5 different morphing methods, including OpenCV (OCV) \cite{openCVmorph}, FaceMorpher (FM) \cite{Facemorpher}, StyleGAN 2 (SG) \cite{DBLP:conf/nips/KarrasAHLLA20,DBLP:conf/iwbf/VenkateshZRRDB20}, and WebMorpher (WM) \cite{webmorph}, and AMSL \cite{amsl}. Each morph method contains 1222 morphed face images generated using only frontal face images with high resolution the dataset also contains 204 BF samples. The FRLL-Morphs is not created or structured to develop MADs \cite{Sarkar2020} and thus does not contain a training set, it is rather developed for the evaluation FR vulnerability to attacks and the MAD performance. Following this, we use this dataset as testing set to evaluate the developed MADs on diverse MAs. Samples of all the used datasets are provided in Fig. \ref{fig:dat_Samples_bf} and \ref{fig:dat_Samples_ma} presenting BF and MA, respectively.


\vspace{-2mm}
\subsection{Evaluation metrics}
\vspace{-1mm}
The MAD performance is presented by the Attack Presentation Classification Error Rate (APCER), i.e. the proportion of MA images incorrectly classified as BF samples, and the BF Presentation Classification Error Rate (BPCER), i.e. the proportion of BF images incorrectly classified as MA samples, as defined in the ISO/IEC 30107-3 \cite{ISO301073}. To cover different operation points, and to present the comparative results, we report the BPCER at three different fixed APCER values (0.1\%, 1.0\%, and 10.0\%). The equal error rate (EER) is also reported, which is the BPCER or APCER value at the decision threshold where they are equal. 
To provide a visual evaluation on a wider operation range, we plot receiver operating characteristic (ROC) curves by plotting the APCER on the x-axis and 1-BPCER on the y-axis at different operational points.
It must be noted again that the MAD evaluation was performed only on the identity-disjoint test data as described in Section \ref{sec:data}.

\begin{figure*}[th!]
    \centering
    \begin{subfigure}[b]{0.46\textwidth}
    \centering
    \includegraphics[width=0.93\linewidth]{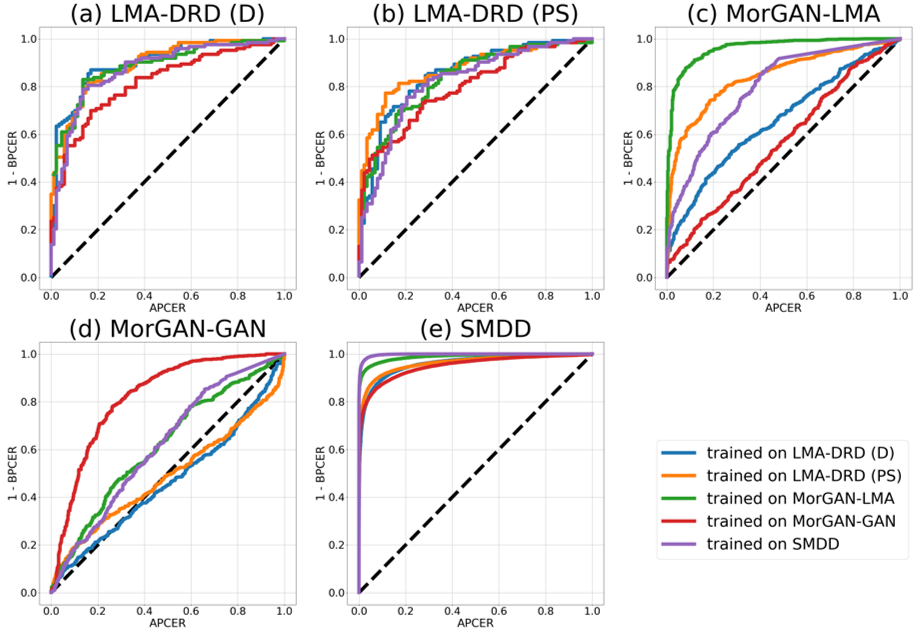}
    \label{figsf1_1}
    \caption{}
    \end{subfigure}
    \begin{subfigure}[b]{0.46\textwidth}
    \centering
    \includegraphics[width=0.93\linewidth]{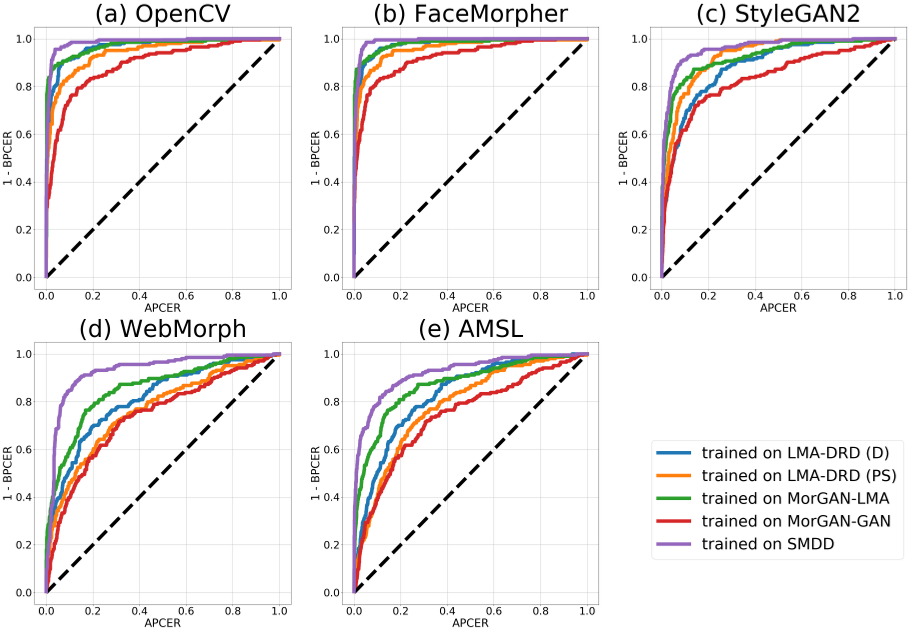}
    \label{figsf1_2}
    \caption{}
    \end{subfigure}
    \vspace{-3mm}
    \caption{ROC curves achieved on the different test sets and different training settings (data) of the MixFaceNets-MAD. This includes intra-dataset and inter-dataset evaluation in (a) and the different morphing techniques of the FRLL-Morphs in (b).}
    \vspace{-3mm}
    \label{fig:ROC_mfn1}
\end{figure*}

\begin{figure*}[th!]
    \centering
    \begin{subfigure}[b]{0.46\textwidth}
    \centering
    \includegraphics[width=0.93\linewidth]{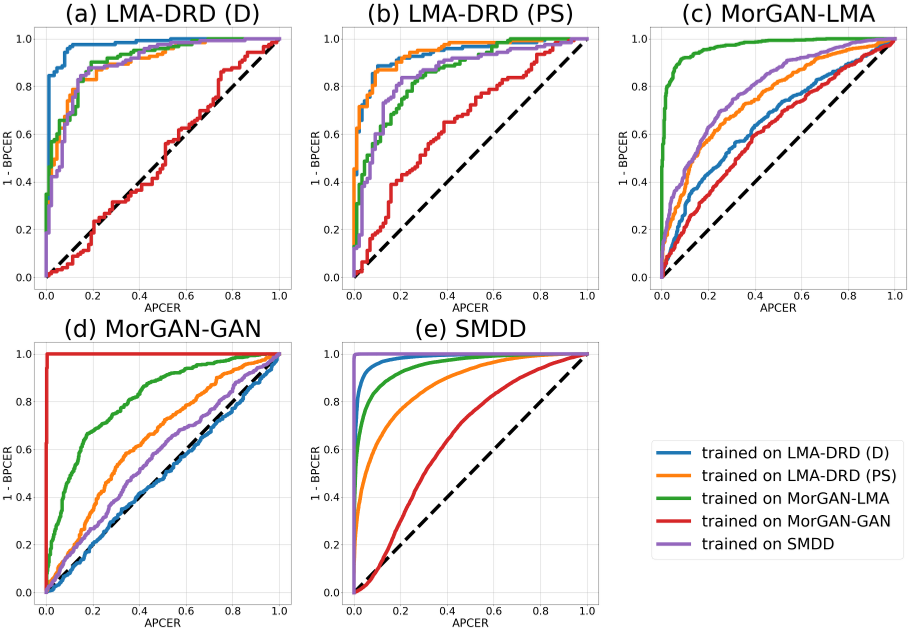}
    \label{figsf2_1}
    \caption{}
    \end{subfigure}
    \begin{subfigure}[b]{0.46\textwidth}
    \centering
    \includegraphics[width=0.93\linewidth]{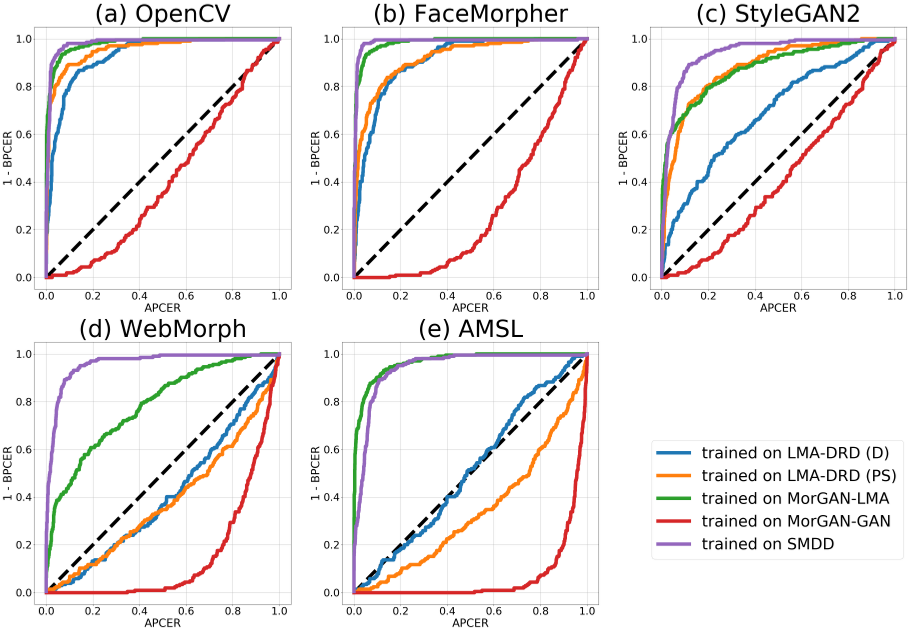}
    \label{figsf2_2}
    \caption{}
    \end{subfigure}
    \vspace{-3mm}
    \caption{ROC curves achieved on the different test sets and different training settings (data) of the Inception-MAD. This includes intra-dataset and inter-dataset evaluation in (a) and the different morphing techniques of the FRLL-Morphs in (b).}
    \vspace{-3mm}
    \label{fig:ROC_Inception1}
\end{figure*}

\begin{figure*}[th!]
    \centering
    \begin{subfigure}[b]{0.46\textwidth}
    \centering
    \includegraphics[width=0.93\linewidth]{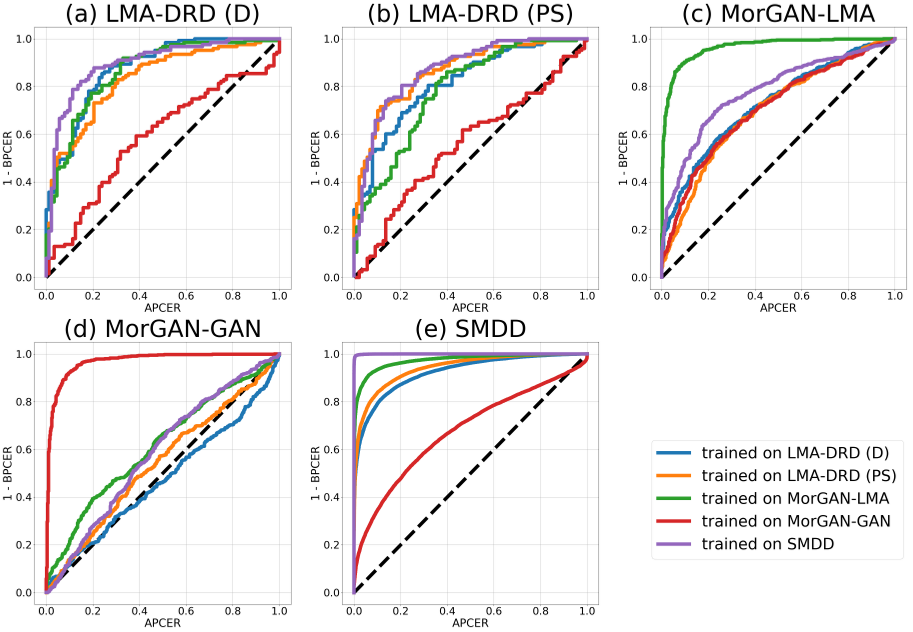}
    \label{figsf3_1}
    \caption{}
    \end{subfigure}
    \begin{subfigure}[b]{0.46\textwidth}
    \centering
    \includegraphics[width=0.93\linewidth]{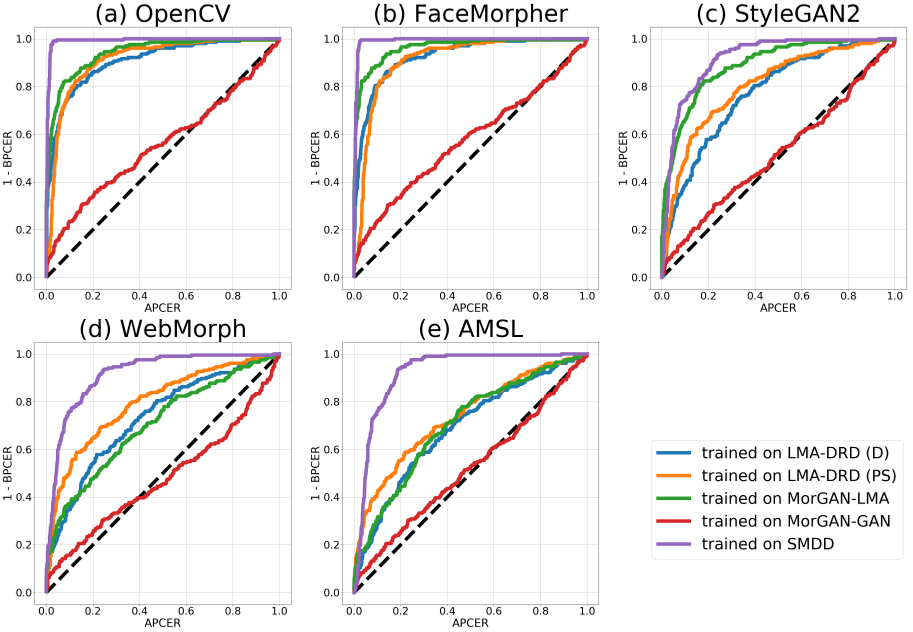}
    \label{figsf3_2}
    \caption{}
    \end{subfigure}
    \vspace{-3mm}
    \caption{ROC curves achieved on the different test sets and different training settings (data) of the PW-MAD. This includes intra-dataset and inter-dataset evaluation in (a) and the different morphing techniques of the FRLL-Morphs in (b).}
    \vspace{-3mm}
    \label{fig:ROC_PW-MAD1}
\end{figure*}

\vspace{-2mm}
\section{Results}
\vspace{-1mm}
The MAD performances on the different MAs of the FRLL-Morphs data achieved by the three MADs trained on the SMDD and the other four training sets are presented in Tab. \ref{tab:res_unknown}.
For all the morphing types in the FRLL-Morphs data, the MADs trained on the SMDD data outperformed the ones trained on other datasets in most of the evaluation settings. Even when the MADs trained on the SMDD did not score the lowest error, they scored a very close error rate to the lowest one. 
In general, FaceMorpher and OpenCV MAs were more detectable than other FRLL-Morphs attacks, scoring EER values below 5\% when the MADs were trained on the SMDD.
While PW-MAD trained on SMDD performed generally better than MicFaceNet-MAD and Inception-MAD on the FaceMorpher and OpenCV MAs, MixFaceNet-MAD performed the best on StyleGAN 2 MAs, and the Inception-MAD performed the best on the WebMorph and AMSL MAs.
Generally, when the training data has a similar morphing strategy as the testing data, e.g. LMA-DRD (D) and FRLL-Morphs OpenCV, the MADs expectedly performed relatively well.

The MAD performances on the test sets of the LMA-DRD (D), LMA-DRD (PS), MorGAN-LMA, MorGAN-GAN, and SMDD, achieved by the three MADs trained on the SMDD and the other four training sets are presented in Tab. \ref{tab:res_known}.
Generally, even though test and train sets are identity-disjoint, the MADs trained on a particular training data achieved high MAD performance on the test set from the same data source, e.g. when MAD trained on MorGAN-LMA training set is evaluated on MorGAN-LMA test set.
When neglecting the unrealistic intra-dataset evaluations, MADs trained on SMDD achieved the best performances (especially with the PW-MAD) on many occasions and were a very close second in many others.
Generally, MorGAN-GAN MAs were the most difficult to detect, however, they pose the least concerns as MAs given the low vulnerability of FR to them and the relatively unrealistic appearance as demonstrated in \cite{DBLP:conf/iwbf/VenkateshZRRDB20}.
In the cases in Tab. \ref{tab:res_known}, the three MADs trained on SMDD achieved comparable performances.
It must be mentioned that the SMDD is not presented as an attack (as it is based on synthetic data) and thus detecting it is not of interest, other than questioning if it can be representative as a privacy-friendly evaluation set. Having this in mind, the SMDD test set did behave similarly (trends in the error rates when different MADs are evaluated) as testing data, compared to the test sets that included attacks created by the same morphing approach, such as LMA-DRD (D) and MorGAN-LMA, however, with significantly lower MAD error rates on the SMDD test set, which might be related to the randomized morph pairing process.

The results in Tab. \ref{tab:res_unknown} and \ref{tab:res_known} are represented over a detailed APCER range as ROC curves in Fig. \ref{fig:ROC_mfn1}, \ref{fig:ROC_Inception1}, \ref{fig:ROC_PW-MAD1}, respectively for the MixFaceNet-MAD, Inception-MAD, and PW-MAD-MAD.
The ROCs approve the discussion driven from Tab. \ref{tab:res_unknown} and \ref{tab:res_known} and assure the applicability of the privacy-friendly and synthetic-based SMDD data as development data for high performing and high generalized MADs. 









\vspace{-2mm}
\section{Conclusion}
\vspace{-1mm}
This work is the first to question the possibility of using synthetic face data as the basis of face MAD development data.
This was driven by presenting an extensive legal analysis on the privacy and security issues raised by using real biometric data.
Towards that, this work presents the publicly available SMDD dataset containing 30,000 attacks and 50,000 bona fide images, all based on synthetic data. 
Experiments on three MAD solutions show the applicability of the SMDD as very effective training data, even when the trained MAD is faced with attacks created with unknown morphing techniques or data sources.

\vspace{-3mm}

\paragraph{Acknowledgment:} This research work has been funded by the German Federal Ministry of Education and Research and the Hessen State Ministry for Higher Education, Research and the Arts within their joint support of the National Research Center for Applied Cybersecurity ATHENE.



{\small
\bibliographystyle{ieee_fullname}
\bibliography{egbib}
}

\end{document}